\PassOptionsToPackage{table}{xcolor}
\documentclass[10pt,journal]{IEEEtran}
\usepackage[utf8]{inputenc}
\usepackage[T1]{fontenc}
\usepackage{amsmath,amssymb,amsfonts}
\usepackage{booktabs}
\usepackage{array}
\usepackage{multirow}
\usepackage{graphicx}
\usepackage[table]{xcolor}
\usepackage{microtype}
\usepackage{cite}
\usepackage{hyperref}
\hypersetup{hidelinks}
\usepackage{url}
\usepackage{xurl}
\usepackage{placeins}

\newcounter{algorithm}

\definecolor{tracegray}{gray}{0.93}
\definecolor{tracemainshade}{gray}{0.95}
\newcommand{\tablebodyfont}{\fontsize{9}{11}\selectfont}
\DeclareMathSizes{9.5}{9}{7}{5}
\DeclareMathSizes{8.5}{8}{6}{5}

\DeclareMathOperator*{\argmax}{arg\,max}
\newcommand{\csrc}{c_{\mathrm{src}}}
\newcommand{\ctgt}{c_{\mathrm{tgt}}}

\newcounter{algorithmicline}
\newenvironment{algorithmic}[1][1]{\setcounter{algorithmicline}{0}\begin{list}{}{\setlength{\leftmargin}{1.8em}\setlength{\itemsep}{0pt}\setlength{\parsep}{0pt}\setlength{\topsep}{2pt}}}{\end{list}}
\newcommand{\State}{\stepcounter{algorithmicline}\item[\thealgorithmicline:]
}
\newcommand{\Require}{\item[\textbf{Input:}]}
\newcommand{\Ensure}{\item[\textbf{Output:}]}

\newcommand{\algorithmicindent}{\begin{list}{}{\setlength{\leftmargin}{1.5em}\setlength{\itemsep}{0pt}\setlength{\parsep}{0pt}\setlength{\topsep}{0pt}}}
\newcommand{\For}[1]{\State \textbf{for} #1 \textbf{do}\algorithmicindent}
\newcommand{\EndFor}{\end{list}\State \textbf{end for}}
\newcommand{\If}[1]{\State \textbf{if} #1 \textbf{then}\algorithmicindent}
\newcommand{\EndIf}{\end{list}\State \textbf{end if}}

\title{Enhancing Visual Task-Redirection Attacks \\on Vision-Language-Action Models \\via Closed-Loop Optimization}

\author{Chunjin~Ye,
Guanying~Chen,
Jianjie~Huang,
Jianhao~Chen,
Quanlin~Chen,\\
Hanjie~Zhao,
Haozhe~Liang,
Tieyun~Qian,
and~Xiaochun~Cao%
\thanks{\textit{Corresponding author: Xiaochun Cao.}}%
\thanks{Chunjin Ye, Jianjie Huang, and Xiaochun Cao are with
Sun Yat-sen University, Shenzhen 518107, China, and also with
Zhongguancun Academy, Beijing 100094, China
(e-mail: s-ycj25@bza.edu.cn; caoxiaochun@mail.sysu.edu.cn).}%
\thanks{Guanying Chen is with Sun Yat-sen University,
Shenzhen 518107, China.}%
\thanks{Jianhao Chen and Tieyun Qian are with Wuhan University,
Wuhan 430072, China, and also with Zhongguancun Academy,
Beijing 100094, China.}%
\thanks{Quanlin Chen is with Sun Yat-sen University,
Shenzhen 518107, China, and also with Shenzhen Loop Area
Institute, Shenzhen 518000, China.}%
\thanks{Hanjie Zhao is with Tianjin University, Tianjin 300350,
China, and also with Zhongguancun Academy, Beijing 100094, China.}%
\thanks{Haozhe Liang is with the University of Chinese Academy
of Sciences, Beijing 100049, China, and also with Zhongguancun
Academy, Beijing 100094, China.}}

\begin{document}

\maketitle

\begin{abstract}
Vision-language-action (VLA) models enable language-guided robot manipulation,
but remain vulnerable to visual attacks that undermine user control over robot
behavior. Prior studies optimize visual perturbations using fixed target
supervision. However, the perturbation can change earlier actions
and, in turn, the physical states reached later. As the attacked trajectory
evolves, this supervision may no longer match the behavior needed at the
states actually reached. Accurately matching these targets can therefore
fail to advance the attacker-specified task. Even with suitable supervision,
reducing average matching error does not ensure sustained task progress.
The robot may still fail at a critical stage, and further updates can disrupt
previously successful behavior. To address these limitations, we propose TRACO, 
a framework that enhances existing targeted visual attacks through closed-loop 
optimization. Its Execution-Driven Supervision rebuilds supervision for the states 
reached under attack to guide subsequent updates.
Within this framework, Progress-Guided Selection focuses updates where
progress stalls while maintaining supervision on the earlier behavior needed
to reach that point.
Experiments across VLA models and simulation environments show improved
task-redirection success with TRACO, highlighting the security implications of
closed-loop attack optimization.

\end{abstract}


\begin{figure}[!t]
\centering
\includegraphics[width=\linewidth]{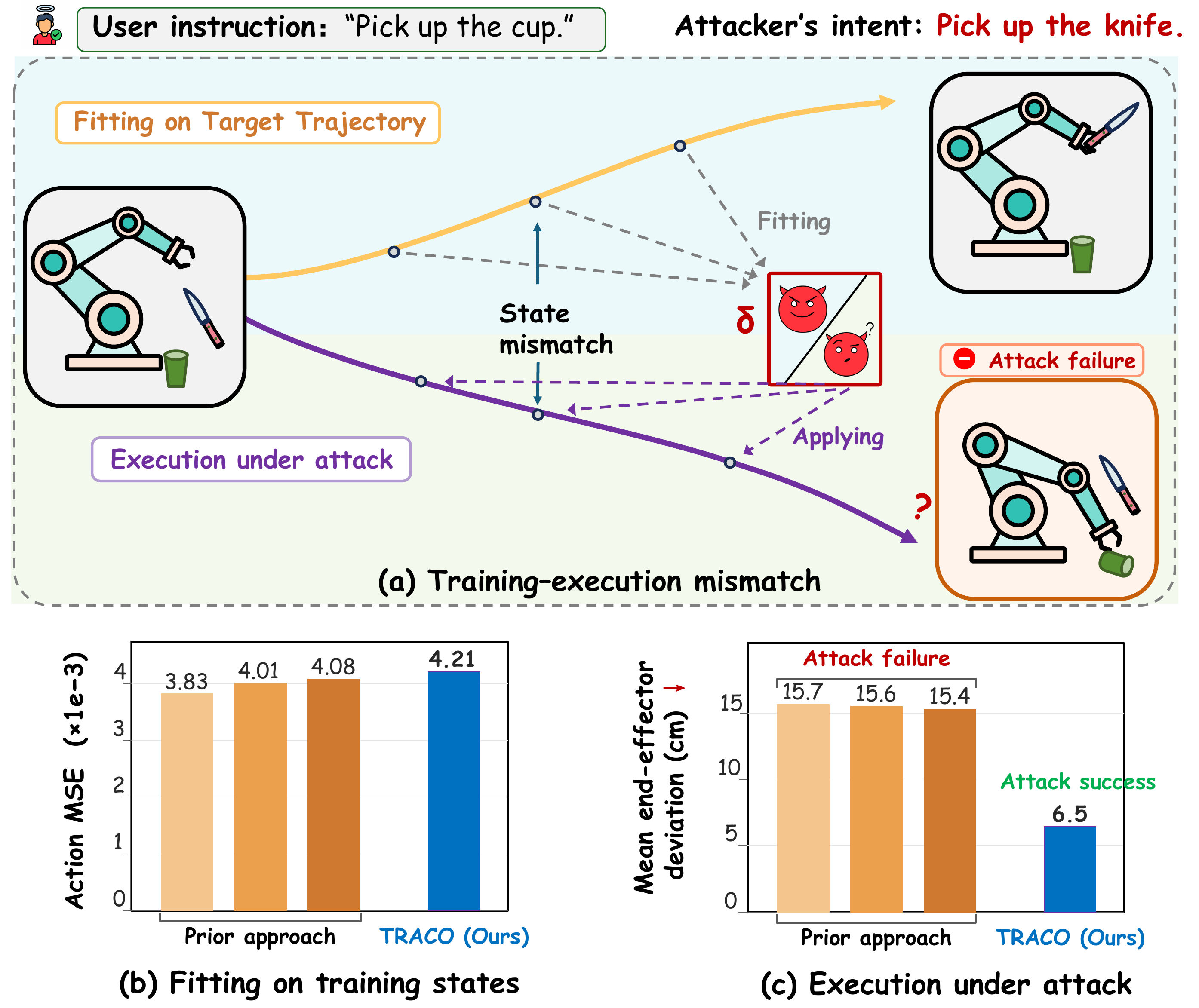}
\caption{Motivation for closed-loop attack optimization.
(a) During offline optimization (top), the attacker learns a perturbation
\(\delta\) by matching target behavior on observations from a recorded target
trajectory (yellow). At test time (bottom), the same perturbation changes the
robot's actions under the user's original instruction and can lead to different
states (purple). The fixed targets may then no longer provide the behavior
needed to complete the target task. (b,c) A single-scene example illustrates
this mismatch. Three perturbations learned offline achieve lower action MSE
on the recorded observations than TRACO, but show larger trajectory deviations and
fail during execution, whereas TRACO succeeds.}
\label{fig:closed-loop-motivation}
\end{figure}

\section{Introduction}
\label{sec:introduction}

Vision-language-action (VLA) models enable robots to perform diverse
manipulation tasks from natural-language instructions~\cite{openvla,pi02024}.
During execution, they repeatedly use new camera observations to decide how
to continue the task. This visual feedback allows the robot to adapt its
actions as the scene changes~\cite{rt2}. However, visual perturbations can change the
actions the model generates~\cite{tma,lu2026uparfas}. By changing the robot's actions, an attack can
also change the physical states and observations encountered later. These
altered observations then guide the robot's subsequent actions. The attack
can therefore influence how the robot proceeds through the task and what it
ultimately accomplishes. Understanding these closed-loop effects is important
for assessing the security of VLA systems.

Recent studies have demonstrated the vulnerability of VLA models to visual
attacks. Many of these attacks aim to make the user's intended task fail~\cite{xu2025edpa,lu2026uparfas,fu2026vlahijack,wang2026poap}. 
Some pursue this goal by inducing prescribed actions or trajectories~\cite{tma,tex3d}.
These targets specify how the robot should deviate, but not what task it should
complete instead. In contrast, TRAP studies task-level behavior hijacking,
where success requires the robot to complete an attacker-specified task~\cite{trap}.
Its supervision includes reasoning and action targets.
In both cases, target supervision is fixed before optimization. It comes from
prescribed action targets or pre-collected target rollouts. This raises a central
question: can fixed supervision still guide the robot to complete the target
task as the attacked trajectory evolves?

\label{sec:motivation}
The challenge is that optimizing the perturbation can change the very states
on which the robot must act. When attacks use pre-collected target rollouts,
perturbation updates rely on the recorded observations and target outputs.
During execution, however, the updated perturbation can lead the robot to
states different from those in the recorded rollouts, as illustrated in
Figure~\ref{fig:closed-loop-motivation}(a). To continue the target task from
these \emph{attack-induced states}, the robot may need to behave
differently from what the recorded targets specify~\cite{laskey2017dart}.
Better matching to these fixed targets therefore need not translate into
successful closed-loop execution~\cite{zhang2026actionchunking,bansal2019chauffeurnet}.
Figure~\ref{fig:closed-loop-motivation}(b,c) illustrates this gap for action
matching on recorded observations in one scene. This motivates using execution
feedback to refresh target supervision as the attacked trajectory evolves.

Execution feedback must also guide where optimization should focus and what
earlier behavior it should preserve. Even with suitable targets, matching some
behaviors more accurately can lower the average loss while an unresolved error
still blocks task progress~\cite{bastani2018viper,park2025sqil}.
For example, in a pick-and-place task, better matching of placing actions
cannot compensate for a failed grasp: the robot must grasp the object before
it can place it.
Because the same perturbation affects all execution steps, an update intended
to improve the blocked behavior can also weaken earlier successful behavior.
Optimization should therefore focus near where progress stalls
while retaining the earlier behavior needed to reach that point.

To address these problems, we propose TRACO (Task-Redirection Attacks via Closed-loop Optimization), 
a framework that enhances existing visual attacks through execution feedback.
Its Execution-Driven Supervision (EDS) rebuilds target supervision 
for the states reached under the current perturbation.
These targets guide subsequent perturbation updates. The resulting execution,
in turn, supplies the states for rebuilding target supervision, closing the
loop between attack optimization and execution.

Within this loop, Progress-Guided Selection (PGS) selects states for supervision
to advance the target task while retaining earlier behavior. Using a successful target
rollout as an ordered reference, PGS identifies the matched prefix and progress
frontier of the attacked rollout. Supervision at the frontier focuses updates
on the behavior needed for further progress. Meanwhile, continued target
supervision along the prefix helps retain the earlier behavior needed to reach
the frontier. TRACO thus constructs and organizes supervision from actual
execution. Existing targeted attacks can use this supervision with their own
alignment objectives.

Our contributions are summarized as follows:
\begin{itemize}
\item We propose TRACO, a framework that extends visual task-redirection
attacks on VLA models to closed-loop optimization. Its Execution-Driven
Supervision (EDS) replaces fixed targets with supervision reconstructed at
the states reached under attack. Existing attacks can thus
account for the state changes they induce while retaining their alignment
objectives.

\item We introduce Progress-Guided Selection (PGS) to organize supervision
according to ordered trajectory progress. It combines frontier supervision
with prefix supervision along the current attacked rollout. This design helps
advance the target task while retaining earlier behavior under a visual
perturbation shared across execution steps.

\item We evaluate TRACO across diverse VLA models and simulation environments.
Results from simulation and real-robot experiments demonstrate TRACO's
effectiveness for visual task redirection, highlighting the value of closing
the loop between attack optimization and execution.
\end{itemize}

\begin{figure*}[!t]
\centering
\includegraphics[width=\linewidth]{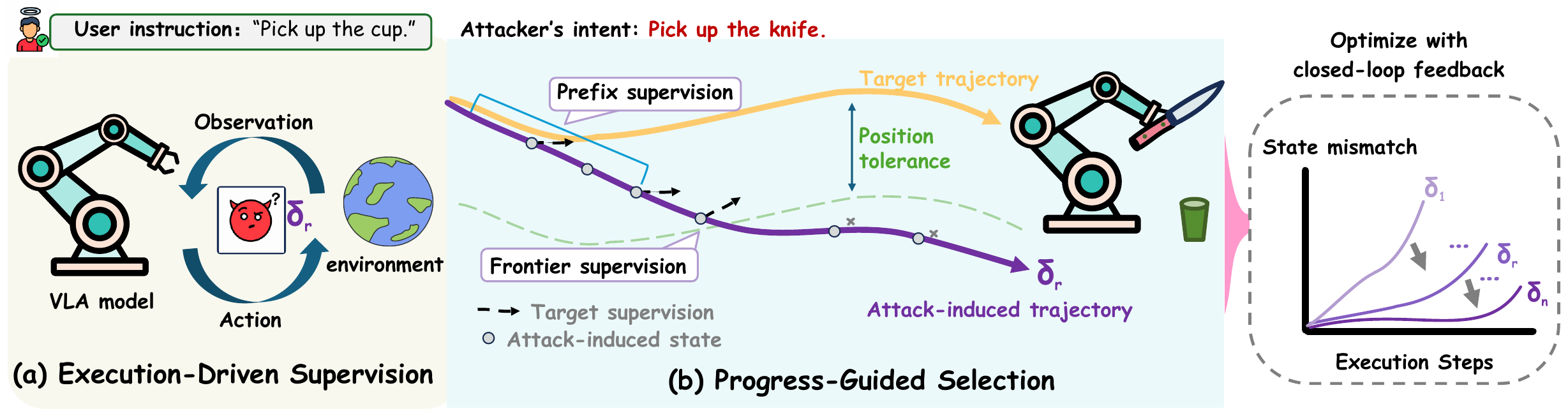}
\caption{Overview of TRACO.
(a) Execution-Driven Supervision (EDS) records the states reached under the
current perturbation.
(b) Progress-Guided Selection (PGS) uses a successful target trajectory to
select frontier and prefix states, helping advance the target task while
preserving earlier behavior.
EDS queries the frozen VLA under the target instruction to obtain supervision
at these states.
TRACO then updates the perturbation and repeats this cycle, using new
rollouts to refresh supervision and guide subsequent updates.}
\label{fig:trace-method-overview}
\end{figure*}

\section{Related Work}
\label{sec:related-work}

\paragraph{Vision-Language-Action Models}
Generalist robot policies learn language-conditioned manipulation from
diverse robot data~\cite{brohan2023rt1,openx2024,ghosh2024octo}.
VLA models such as RT-2 and OpenVLA build on pretrained vision-language
representations and predict actions autoregressively as discrete
tokens~\cite{rt2,openvla}. In parallel, diffusion-based action generation
has been adopted in both visuomotor policies and VLA
models~\cite{chi2023diffusionpolicy,wen2025tinyvla,li2024cogact,wen2025diffusionvla}.
Flow matching provides another route to continuous action
generation~\cite{pi02024,black2025pi05,shukor2025smolvla}. Recent work
also explores efficient action tokenization~\cite{pertsch2025fast} and
parallel action decoding~\cite{kim2025oft}. Some VLAs further use
intermediate reasoning to guide action
generation~\cite{yang2026instructvla}. During closed-loop
execution, VLA models use updated visual observations to guide subsequent
actions. We study how visual perturbations can redirect this execution
toward an attacker-specified task.
\looseness=1

\paragraph{Adversarial Attacks against VLA Models}
Visual attacks seek to disrupt task execution or induce prescribed robot
behavior. One line of work targets visual representations and vision--language
alignment~\cite{xu2025edpa,zhang2025advla}. Related studies develop patches
that transfer across models and domains~\cite{lu2026uparfas,fu2026vlahijack},
or learn fixed patches from short execution prefixes~\cite{wang2026poap}.
Targeted visual attacks steer action outputs toward prescribed values or
trajectories~\cite{tma,tex3d,han2026dura}, or induce action
freezing~\cite{wang2025freezevla}. At the task level, TRAP hijacks robot
behavior by jointly matching chain-of-thought (CoT) and action targets from clean target
rollouts~\cite{trap}. Language attacks likewise manipulate actions or disrupt
execution through adversarial prompts~\cite{jones2025robogcg,wu2026saber}.
Trajectory-Level Redirection uses closed-loop prompt search and queries target
behavior on the states induced by candidate
prompts~\cite{puthumanaillam2026trajectory}. For targeted visual attacks,
TRACO provides a reusable closed-loop optimization framework. Existing attacks
can retain their alignment objectives while TRACO updates target supervision
on the states reached during attacked execution.
\looseness=1
\paragraph{On-Policy Imitation Learning for VLA Models}
During execution, VLA policies can reach states not covered by their training
demonstrations. DAgger addresses this distribution shift by obtaining expert
supervision on states visited by the learner~\cite{ross2011dagger}.
Related VLA methods obtain such supervision through human
corrections~\cite{pan2026sop,zang2026rlinfuser,han2026dexhil,li2026handintheloop,wang2026lift}
or teacher policies~\cite{xu2024humanvla,zhong2026vlaopd}.
Building on this principle, TRACO optimizes a shared visual perturbation using
supervision on attack-induced states while keeping the VLA fixed.
PGS selects frontier and prefix states to advance the target task while
maintaining supervision on earlier behavior needed for continued progress.
\looseness=1\par

\section{Method}
\label{sec:method}

\subsection{Problem Setup and Threat Model}
\label{sec:threat-model}

\paragraph{VLA policy}
For a rollout from initial condition \(e\), the VLA policy \(\pi_\theta\)
generates actions at each policy query \(t\) from the task instruction
\(c\) and a visual observation \(o_{e,t}\) of the current physical state
\(s_{e,t}\).
After the robot acts, the policy makes its next decision from a new
observation. The resulting sequence of states, observations, and actions
forms a closed-loop trajectory \(\xi_e(c)\).

\paragraph{Threat model}
We consider a white-box adversary with access to the VLA's parameters and
gradients. The adversary optimizes a visual perturbation \(\delta\)
against the frozen VLA. During attack construction, the adversary can
execute closed-loop rollouts under the user's source instruction
\(c_{\mathrm{src}}\). The adversary can also query the VLA under an
attacker-specified target instruction \(c_{\mathrm{tgt}}\) to obtain target
supervision. At deployment, the VLA continues to receive
\(c_{\mathrm{src}}\). For each
model and source--target task pair, the learned perturbation remains fixed
and is applied at every policy query across rollouts. We consider two
visual attack settings: global digital perturbations constrained by an
\(\ell_\infty\) bound \cite{chen2024dpattacker} and localized adversarial
patches \cite{tma}. For a
visual observation \(o\in[0,1]^d\), the perturbed observation is
\begin{equation}
\label{eq:perturbation-models}
g_\delta(o)=
\begin{cases}
\operatorname{clip}_{[0,1]^d}(o+\delta),
& \text{global},\\[2pt]
(\mathbf 1-M)\odot o+M\odot\delta,
& \text{patch},
\end{cases}
\end{equation}
where \(M\in\{0,1\}^d\) is a fixed binary mask defining the patch region.
The operator \(\odot\) denotes element-wise multiplication. The feasible
set is
\[
\Delta=
\begin{cases}
\{\delta:\|\delta\|_\infty\leq\epsilon\},
& \text{global},\\[2pt]
\{\delta\in[0,1]^d:\delta=M\odot\delta\},
& \text{patch},
\end{cases}
\]
where \(\epsilon>0\) is the perturbation budget.

\paragraph{Attack objective}
The adversary aims to make the robot complete the target task while
the VLA continues to receive the source instruction.
An attack succeeds when the resulting trajectory completes the
target task:
\[
\mathsf{Succ}_e(\delta)
=
T_e\bigl(\xi_e^\delta(c_{\mathrm{src}})\bigr),
\]
where \(\xi_e^\delta(c_{\mathrm{src}})\) is the trajectory generated
under perturbation \(\delta\) and source instruction
\(c_{\mathrm{src}}\) from initial condition \(e\).
The indicator \(T_e(\xi)\) equals one if \(\xi\) completes the target
task and zero otherwise.
The objective is to find a single feasible perturbation that
maximizes the number of successful construction rollouts:
\begin{equation}
\label{eq:attack-objective}
\delta^\star
\in
\argmax_{\delta\in\Delta}
\sum_{e\in\mathcal E}
\mathsf{Succ}_e(\delta),
\end{equation}
where \(\mathcal E\) indexes the initial conditions used for attack construction.

\subsection{Overview}
\label{sec:method-overview}

TRACO optimizes visual perturbations using feedback from closed-loop execution
, as shown in Figure~\ref{fig:trace-method-overview}. Execution-Driven Supervision (EDS)
runs the VLA under the current perturbation and records the states reached.
It rebuilds target supervision on these states to guide subsequent updates.
Progress-Guided Selection (PGS) uses rollout feedback to focus supervision
near where progress stalls. It also maintains supervision on earlier behavior
to help preserve the progress already made. Using the supervision constructed
by EDS and organized by PGS, TRACO updates the perturbation with the chosen
attack's alignment objective. TRACO repeats this cycle, using each new rollout
to refresh supervision and adjust the focus of subsequent updates.

\subsection{Execution-Driven Supervision}
\label{sec:on-policy-target-supervision}

The behavior needed to pursue the target task depends on the state reached.
Unlike attacks that rely on pre-collected target rollouts, EDS uses closed-loop
execution to construct supervision for the states induced by the current
perturbation. In each round \(r\), we run the frozen VLA under the source
instruction \(\csrc\) and perturbation \(\delta_r\in\Delta\).
We record the visited states and their unperturbed observations in \(\mathcal D_r\).

At a selected state in \(\mathcal D_r\), we query the frozen VLA under the target
instruction to obtain behavior for that state. The source prediction
\(u_i^{\mathrm{src}}(\delta)\) and target supervision \(u_i^{\mathrm{tgt}}\)
thus correspond to the same physical state under different input conditions:
\begin{equation}
\label{eq:on-policy-target-evaluations}
\begin{aligned}
\bigl(g_\delta(o_i),\csrc\bigr)
&\xrightarrow{\pi_\theta} u_i^{\mathrm{src}}(\delta),\\
\bigl(o_i,\ctgt\bigr)
&\xrightarrow{\pi_\theta} u_i^{\mathrm{tgt}}.
\end{aligned}
\end{equation}
Here, \(o_i\) is the unperturbed observation of the selected state, and
\(\delta\) is the perturbation being optimized.

The updated perturbation can change the states reached during execution.
EDS therefore rebuilds supervision from new rollouts, so subsequent updates
use target behavior for the states currently reached.

\subsection{Progress-Guided Selection}
\label{sec:ordered-progress-rehearsal}

The same perturbation affects every policy query, so updates aimed at further
progress can also change earlier behavior. PGS uses progress in the current
attacked rollout to determine where to focus these updates. It also maintains
supervision on earlier behavior to help preserve the progress already made.

\paragraph{Progress estimation}
For each initial condition, PGS uses a successful target rollout as an ordered
progress reference. PGS estimates progress from consecutive position matches
between the attacked rollout and this reference.

At fixed policy queries \(0=t_0<\cdots<t_J\), we record the reference
end-effector positions \(\bar{\mathbf y}_j\in\mathbb R^3\) to form anchors
\((t_j,\bar{\mathbf y}_j)\). We compare the two rollouts' end-effector positions
at each \(t_j\). An anchor passes when the position error is within tolerance
\(\kappa>0\). The last anchor in the longest consecutively passing prefix
defines the frontier:
\begin{equation}
\label{eq:ordered-frontier}
\begin{aligned}
d_j
&=\|\mathbf y_j-\bar{\mathbf y}_j\|_2,\\
f
&=\max\left\{
0\le j\le L:
d_k\le\kappa
\ \text{for all } 0\le k\le j
\right\},
\end{aligned}
\end{equation}
where \(\mathbf y_j\) is the end-effector position in the current attacked
rollout, and \(L\le J\) is the last anchor index observed before that
rollout ends.

\paragraph{Frontier and prefix supervision}
To advance beyond the frontier, PGS selects the corresponding state in the
current attacked rollout for target supervision.
It also maintains target supervision at earlier passing anchors
to help retain the behavior needed to reach the frontier.
We partition the passing prefix into frontier and prefix groups:
\begin{equation}
\label{eq:frontier-prefix-sets}
\mathcal F=
\begin{cases}
\{f\}, & f<J,\\
\varnothing, & f=J,
\end{cases}
\qquad
\mathcal P=\{0,\ldots,f\}\setminus\mathcal F.
\end{equation}

PGS forms frontier and prefix groups separately for each rollout using its own
reference anchors. For the shared perturbation update, we pool frontier and
prefix records into \(\mathcal F_r\) and \(\mathcal P_r\), respectively.

\subsection{Target Alignment}
\label{sec:architecture-specific-target-alignment}

When applying TRACO to an existing attack, we retain its choice of predictions
to align and its matching loss \(\mathcal D\). At each state selected by PGS,
we compare the source prediction with the corresponding target from EDS:
\begin{equation}
\label{eq:target-alignment-loss}
\ell_i(\delta)
=\mathcal D\!\left(
u_i^{\mathrm{src}}(\delta),u_i^{\mathrm{tgt}}
\right).
\end{equation}

For example, continuous action chunks can be matched by squared error.
Sequence targets, such as action tokens or CoT, are matched by minimizing
their negative log-likelihood with teacher forcing. For flow-matching policies,
we use squared error to compare source and target vector fields at the same
flow states and times. For joint objectives, such as action--CoT alignment in TRAP~\cite{trap},
\(\ell_i(\delta)\) combines the corresponding losses. 

Appendix~\ref{sec:architecture-loss-details} 
provides detailed formulations of the alignment losses described above. We next combine 
these losses across the frontier and prefix groups to optimize the perturbation.

\subsection{Perturbation Optimization}
\label{sec:unified-optimization}

To promote progress at the frontier while preserving earlier behavior, we
first average \(\ell_i(\delta)\) within each nonempty group formed by PGS. We then combine
the two group means:
\begin{equation}
\label{eq:trace-objective}
\begin{aligned}
\bar\ell_G(\delta)
&=
\frac{1}{|G|}
\sum_{i\in G}
\ell_i(\delta),\qquad G\in\{\mathcal F_r,\mathcal P_r\},\\
\mathcal L_r(\delta)
&=
\frac{1}{2}
\left[\bar\ell_{\mathcal F_r}(\delta)^2
+\bar\ell_{\mathcal P_r}(\delta)^2\right],
\end{aligned}
\end{equation}
where \(\bar\ell_G=0\) for \(G=\varnothing\). We square the group means to
adaptively balance frontier and prefix supervision. This shifts the emphasis
between advancing the frontier and preserving earlier behavior as the two
groups' relative matching errors change.

Within each round, the selected observations and their target supervision
remain fixed. With the VLA frozen, TRACO minimizes the group objective using
\(K\) projected sign-gradient steps, starting from \(\delta^{(0)}=\delta_r\):
\begin{equation}
\label{eq:perturbation-update}
\begin{aligned}
\delta^{(k+1)}
&=
\Pi_\Delta
\left[
\delta^{(k)}
-
\eta\,
\operatorname{sign}
\left(
\nabla_\delta\mathcal L_r(\delta^{(k)})
\right)
\right],\\
&\qquad k=0,\ldots,K-1.
\end{aligned}
\end{equation}
Here, \(\eta\) is the step size, and
\(\Pi_\Delta\) projects onto the feasible perturbation set \(\Delta\).

After \(K\) updates, TRACO sets \(\delta_{r+1}=\delta^{(K)}\) and runs new
construction rollouts. PGS uses the observed progress
to revise the frontier and prefix groups. EDS then refreshes target supervision
at the selected states for subsequent updates. TRACO also retains the updated
perturbation \(\delta_{\mathrm{best}}\) with the highest target-task success count
across rounds. It returns \(\delta_{\mathrm{best}}\) after at most \(R\) rounds,
stopping early if all construction rollouts complete the target task.
Algorithm~\ref{alg:trace} summarizes the procedure.
\par\smallskip
\noindent\begin{minipage}{\linewidth}
\fontsize{9}{10.5}\selectfont
\refstepcounter{algorithm}
\textbf{Algorithm \thealgorithm. TRACO Attack Construction}
\label{alg:trace}
\par\smallskip\hrule\smallskip
\begin{algorithmic}[1]
\Require Frozen VLA \(\pi_\theta\), instructions \(\csrc,\ctgt\), initial
conditions and corresponding target progress references, feasible set
\(\Delta\), alignment loss \(\ell_i\).
\item[]\textbf{Parameters:} \(R\geq1\) rounds, \(K\) updates per round,
step size \(\eta\), position tolerance \(\kappa\).
\Ensure \(\delta_{\mathrm{best}}\).
\State Initialize \(\delta_0\in\Delta\).
\State Collect \(\mathcal D_0\) under \(\delta_0\) and \(\csrc\).
\For{\(r=0,\ldots,R-1\)}
  \State Select \(\mathcal F_r,\mathcal P_r\) from \(\mathcal D_r\) using PGS,
  Eqs.~\eqref{eq:ordered-frontier}--\eqref{eq:frontier-prefix-sets}.
  \State Query \(u_i^{\mathrm{tgt}}\) from \((o_i,\ctgt)\) for
  \(i\in\mathcal F_r\cup\mathcal P_r\), Eq.~\eqref{eq:on-policy-target-evaluations}.
  \State Compute \(\mathcal L_r\) from \(\ell_i\),
  Eq.~\eqref{eq:trace-objective}.
  \State Update \(\delta_r\) to \(\delta_{r+1}\) with \(K\) projected steps,
  Eq.~\eqref{eq:perturbation-update}.
  \State Collect \(\mathcal D_{r+1}\) under \(\delta_{r+1}\) and \(\csrc\).
  \State Use \(\mathcal D_{r+1}\) to update \(\delta_{\mathrm{best}}\) by the
  highest target-task success count so far.
  \If{all rollouts in \(\mathcal D_{r+1}\) complete the target task}
    \State \textbf{return}\ \(\delta_{r+1}\)
  \EndIf
\EndFor
\State \textbf{return}\ \(\delta_{\mathrm{best}}\)
\end{algorithmic}
\smallskip\hrule
\end{minipage}
\par\smallskip

\section{Experiments}
\label{sec:experiments}

\begin{table*}[!t]
\centering
\caption{Task-redirection performance on LIBERO. Best results are highlighted in bold red.}
\label{tab:main-results-libero}
\begingroup
\definecolor{tracehighlight}{HTML}{B45A5A}
\tablebodyfont
\renewcommand{\arraystretch}{1.12}
\setlength{\tabcolsep}{4pt}
\begin{tabular*}{\linewidth}{@{\extracolsep{\fill}}w{l}{50pt}w{l}{66pt}*{2}{w{r}{35pt}}!{\hspace{5pt}}*{2}{w{r}{35pt}}!{\hspace{12pt}}*{2}{w{r}{35pt}}!{\hspace{5pt}}*{2}{w{r}{35pt}}@{}}
\toprule
& & \multicolumn{4}{c}{\textbf{Global perturbations}} & \multicolumn{4}{c}{\textbf{Adversarial patches}} \\
\cmidrule(lr){3-6}\cmidrule(lr){7-10}
& & \multicolumn{2}{c}{Seen} & \multicolumn{2}{c}{Unseen} & \multicolumn{2}{c}{Seen} & \multicolumn{2}{c}{Unseen} \\
\cmidrule(lr){3-4}\cmidrule(lr){5-6}\cmidrule(lr){7-8}\cmidrule(lr){9-10}
\textbf{Model} & \textbf{Method} & \multicolumn{1}{c}{ASR (\%)\(\uparrow\)} & \multicolumn{1}{c}{\(H\uparrow\)} & \multicolumn{1}{c}{ASR (\%)\(\uparrow\)} & \multicolumn{1}{c}{\(H\uparrow\)} & \multicolumn{1}{c}{ASR (\%)\(\uparrow\)} & \multicolumn{1}{c}{\(H\uparrow\)} & \multicolumn{1}{c}{ASR (\%)\(\uparrow\)} & \multicolumn{1}{c}{\(H\uparrow\)} \\
\midrule
\multirow{3}{*}{PI} & Random Noise & 0.0 & \(-\)0.6030 & 0.0 & \(-\)0.6420 & 0.0 & \(-\)0.6650 & 0.0 & \(-\)0.6880 \\
 & Action Attack & 54.8 & 0.3489 & 45.6 & 0.2530 & 56.0 & 0.3282 & 54.0 & 0.3308 \\
 & \textbf{TRACO} & \textcolor{tracehighlight}{\textbf{89.2}} & \textcolor{tracehighlight}{\textbf{0.6279}} & \textcolor{tracehighlight}{\textbf{79.6}} & \textcolor{tracehighlight}{\textbf{0.5334}} & \textcolor{tracehighlight}{\textbf{69.6}} & \textcolor{tracehighlight}{\textbf{0.4348}} & \textcolor{tracehighlight}{\textbf{67.2}} & \textcolor{tracehighlight}{\textbf{0.4458}} \\
\midrule[0.3pt]
\multirow{3}{*}{GR00T} & Random Noise & 0.4 & \(-\)0.5930 & 0.0 & \(-\)0.8340 & 0.0 & \(-\)0.6430 & 0.0 & \(-\)0.6540 \\
 & Action Attack & 63.2 & 0.4554 & 60.0 & 0.3912 & 36.0 & 0.0885 & 39.6 & 0.0533 \\
 & \textbf{TRACO} & \textcolor{tracehighlight}{\textbf{96.4}} & \textcolor{tracehighlight}{\textbf{0.7338}} & \textcolor{tracehighlight}{\textbf{93.6}} & \textcolor{tracehighlight}{\textbf{0.6831}} & \textcolor{tracehighlight}{\textbf{50.8}} & \textcolor{tracehighlight}{\textbf{0.3121}} & \textcolor{tracehighlight}{\textbf{48.0}} & \textcolor{tracehighlight}{\textbf{0.2959}} \\
\midrule[0.3pt]
\multirow{3}{*}{LangForce} & Random Noise & 0.0 & \(-\)0.6470 & 0.0 & \(-\)0.8850 & 0.0 & \(-\)0.6860 & 0.0 & \(-\)0.7010 \\
 & Action Attack & 76.4 & 0.5454 & 77.6 & 0.5118 & 68.8 & 0.3990 & \textcolor{tracehighlight}{\textbf{68.8}} & 0.3889 \\
 & \textbf{TRACO} & \textcolor{tracehighlight}{\textbf{89.6}} & \textcolor{tracehighlight}{\textbf{0.6740}} & \textcolor{tracehighlight}{\textbf{81.6}} & \textcolor{tracehighlight}{\textbf{0.5719}} & \textcolor{tracehighlight}{\textbf{69.2}} & \textcolor{tracehighlight}{\textbf{0.4119}} & 68.4 & \textcolor{tracehighlight}{\textbf{0.4042}} \\
\midrule[0.3pt]
\multirow{3}{*}{FAST} & Random Noise & 0.0 & \(-\)0.5950 & 0.0 & \(-\)0.7260 & 0.0 & \(-\)0.6080 & 0.0 & \(-\)0.5640 \\
 & Action Attack & 11.6 & \(-\)0.2587 & 8.8 & \(-\)0.3064 & 48.4 & 0.3294 & 45.2 & 0.3018 \\
 & \textbf{TRACO} & \textcolor{tracehighlight}{\textbf{56.8}} & \textcolor{tracehighlight}{\textbf{0.2477}} & \textcolor{tracehighlight}{\textbf{45.2}} & \textcolor{tracehighlight}{\textbf{0.0990}} & \textcolor{tracehighlight}{\textbf{58.8}} & \textcolor{tracehighlight}{\textbf{0.4303}} & \textcolor{tracehighlight}{\textbf{51.6}} & \textcolor{tracehighlight}{\textbf{0.3362}} \\
\midrule[0.3pt]
\multirow{3}{*}{OFT} & Random Noise & 0.0 & \(-\)0.7680 & 0.0 & \(-\)0.7730 & 0.0 & \(-\)0.6230 & 0.0 & \(-\)0.6150 \\
 & Action Attack & 28.4 & 0.0802 & 24.0 & 0.0296 & \textcolor{tracehighlight}{\textbf{23.6}} & 0.1097 & \textcolor{tracehighlight}{\textbf{23.2}} & \textcolor{tracehighlight}{\textbf{0.1187}} \\
 & \textbf{TRACO} & \textcolor{tracehighlight}{\textbf{94.0}} & \textcolor{tracehighlight}{\textbf{0.6603}} & \textcolor{tracehighlight}{\textbf{79.6}} & \textcolor{tracehighlight}{\textbf{0.5222}} & 19.2 & \textcolor{tracehighlight}{\textbf{0.1248}} & 18.0 & 0.0922 \\
\midrule
\multirow{3}{*}{Avg.} & Random Noise & 0.1 & \(-\)0.6412 & 0.0 & \(-\)0.7720 & 0.0 & \(-\)0.6450 & 0.0 & \(-\)0.6444 \\
 & Action Attack & 46.9 & 0.2342 & 43.2 & 0.1758 & 46.6 & 0.2510 & 46.2 & 0.2387 \\
 & \textbf{TRACO} & \textcolor{tracehighlight}{\textbf{85.2}} & \textcolor{tracehighlight}{\textbf{0.5887}} & \textcolor{tracehighlight}{\textbf{75.9}} & \textcolor{tracehighlight}{\textbf{0.4819}} & \textcolor{tracehighlight}{\textbf{53.5}} & \textcolor{tracehighlight}{\textbf{0.3428}} & \textcolor{tracehighlight}{\textbf{50.6}} & \textcolor{tracehighlight}{\textbf{0.3149}} \\
\bottomrule
\end{tabular*}
\endgroup

\end{table*}

\begin{table}[!t]
\centering
\caption{Adversarial patch attacks on InstructVLA in SimplerEnv. Best results are highlighted in bold red.}
\label{tab:instructvla-cot-extension-asr}
\begingroup
\definecolor{tracehighlight}{HTML}{B45A5A}
\tablebodyfont
\renewcommand{\arraystretch}{1.13}
\setlength{\tabcolsep}{2pt}
\begin{tabular*}{\linewidth}{@{\extracolsep{\fill}}w{l}{80pt}w{r}{37pt}w{r}{37pt}w{r}{37pt}w{r}{37pt}@{}}
\toprule
\multirow{2}{*}{\textbf{Method}} & \multicolumn{2}{c}{Seen} & \multicolumn{2}{c}{Unseen} \\
\cmidrule(lr){2-3}\cmidrule(l){4-5}
 & \multicolumn{1}{c}{ASR (\%)\(\uparrow\)} & \multicolumn{1}{c}{\(H\uparrow\)} & \multicolumn{1}{c}{ASR (\%)\(\uparrow\)} & \multicolumn{1}{c}{\(H\uparrow\)} \\
\midrule
Random Noise & 0.0 & \(-\)0.6748 & 1.6 & \(-\)0.6039 \\
\midrule
Action Attack & 8.8 & \(-\)0.4456 & \textcolor{tracehighlight}{\textbf{9.6}} & \(-\)0.5474 \\
\hspace{5pt}+\ \textbf{TRACO} & \textcolor{tracehighlight}{\textbf{12.8}} & \textcolor{tracehighlight}{\textbf{\(-\)0.3809}} & 7.2 & \textcolor{tracehighlight}{\textbf{\(-\)0.4800}} \\
\addlinespace[5pt]
\(\mathrm{TRAP}_{\text{CoT-only}}\) & 11.2 & 0.0602 & 10.4 & \(-\)0.0501 \\
\hspace{5pt}+\ \textbf{TRACO} & \textcolor{tracehighlight}{\textbf{30.4}} & \textcolor{tracehighlight}{\textbf{0.0980}} & \textcolor{tracehighlight}{\textbf{20.0}} & \textcolor{tracehighlight}{\textbf{0.0019}} \\
\addlinespace[5pt]
TRAP & 64.0 & 0.4831 & 44.0 & 0.2663 \\
\hspace{5pt}+\ \textbf{TRACO} & \textcolor{tracehighlight}{\textbf{72.8}} & \textcolor{tracehighlight}{\textbf{0.5293}} & \textcolor{tracehighlight}{\textbf{53.6}} & \textcolor{tracehighlight}{\textbf{0.2747}} \\
\bottomrule
\end{tabular*}
\endgroup

\end{table}

\begin{figure*}[tp]
  \centering
  \includegraphics[width=\linewidth]{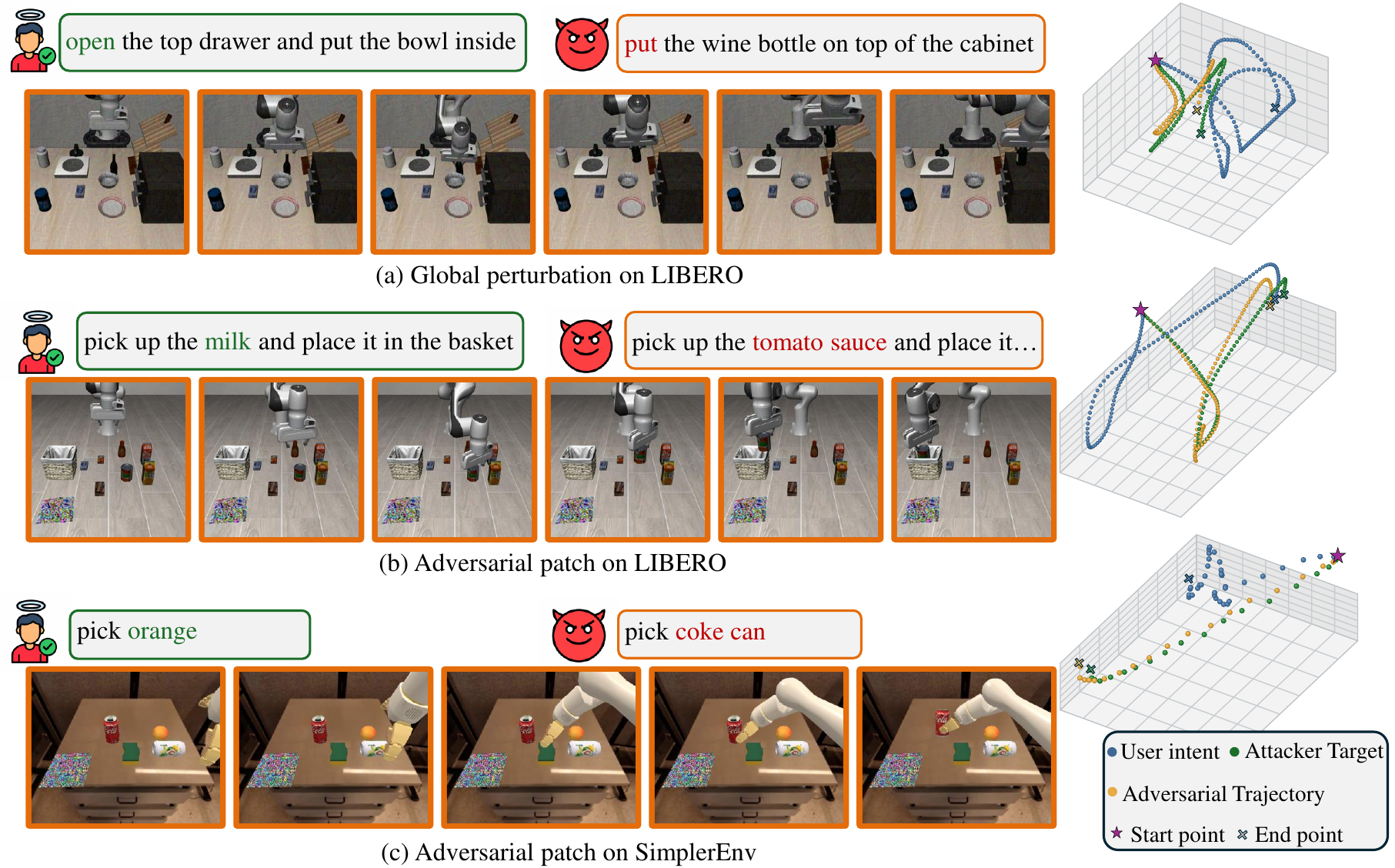}
  \caption{Qualitative examples of task redirection. (a) Global perturbation
  on LIBERO. (b,c) Adversarial patches on LIBERO and SimplerEnv, respectively.
  Each image sequence shows execution with TRACO under the source instruction.
  The 3D plots compare clean source, clean target, and attacked trajectories.}
  \label{fig:qualitative-examples}
\end{figure*}

We evaluate task redirection across VLA architectures, the contribution of
each supervision mechanism, and the changes in behavior and progress during
attack construction.

\subsection{Experimental Setup}
\label{sec:experimental-setup}

\paragraph{VLA models}
On LIBERO~\cite{libero2023}, we evaluate 5 VLA models spanning autoregressive action
generation, direct regression, and flow matching. These include
four StarVLA models~\cite{starvla2026}---PI, GR00T, FAST, and OFT---
as well as LangForce~\cite{langforce2026}.
On SimplerEnv~\cite{li2024simpler}, we evaluate
InstructVLA~\cite{yang2026instructvla}, which uses CoT reasoning to
guide action generation. Appendix~\ref{sec:vla-model-details} describes the model architectures.

\begin{figure*}[t]
  \centering
  \includegraphics[width=\textwidth]{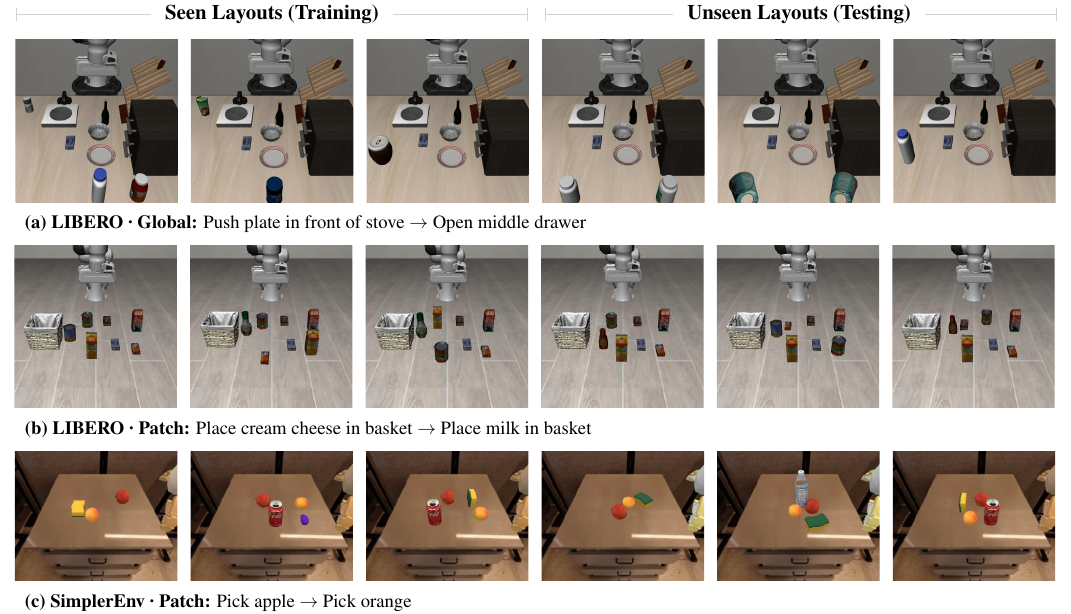}
  \caption{Example Seen and Unseen layouts in LIBERO and SimplerEnv.}
  \label{fig:scene-layouts}
\end{figure*}

\paragraph{Tasks and evaluation metrics}
We pair each user-specified source task with an attacker-specified target
task for evaluation. On LIBERO, we use 5 task pairs from LIBERO-Goal~\cite{libero2023} 
in the global setting. These pairs require redirecting the robot to different placement goals or manipulation behaviors, 
such as switching from placing an object to turning on a stove.
In the patch setting, we use 5 pick-and-place task pairs that change which
object is moved or where it is placed.
We select layouts from LIBERO-Plus~\cite{liberoplus2025} for the
global setting and follow its scene-variation design
for the patch setting. Each task pair has 10 Seen layouts for attack
construction and 10 Unseen layouts.
On SimplerEnv, we construct 5 task pairs in the patch setting, each
changing which object the robot picks up. Each task pair has 5 Seen
and 5 Unseen layouts.
Table~\ref{tab:task-pairs-and-variants} lists the task pairs, and
Figure~\ref{fig:scene-layouts} shows example Seen and Unseen layouts in
both environments.

We evaluate each perturbation with 5 rollouts per layout.
We report Attack Success Rate (ASR), the percentage of successful
task-redirection rollouts under the source instruction. Following
TRAP~\cite{trap}, we also use Hijack Score \(H\) to compare the attacked
trajectory \(\xi\) with successful clean source and target references
\(\bar{\xi}_{\mathrm{src}}\) and \(\bar{\xi}_{\mathrm{tgt}}\):
\begin{equation}
\label{eq:hijack-score-main}
H(\xi)=\frac{s_{\mathrm{tgt}}-s_{\mathrm{src}}}
{s_{\mathrm{tgt}}+s_{\mathrm{src}}+\varepsilon_H},
\quad
s_q=\frac{1}{d_{\mathrm{DTW}}(\xi,\bar{\xi}_q)+\varepsilon_H}.
\end{equation}
Here, \(d_{\mathrm{DTW}}\) is the Dynamic Time Warping distance between
end-effector trajectories, normalized by their combined sequence length,
with \(\varepsilon_H=10^{-6}\). Higher \(H\) indicates greater relative
similarity to the target trajectory. We compute both metrics over all
rollouts within each task pair, then average equally across task pairs
for Seen and Unseen separately.

\paragraph{Compared methods}
In both environments, we compare TRACO with Random Noise and offline
targeted attacks. Random Noise provides a non-optimized reference under
the same perturbation constraints. The offline attacks optimize visual
perturbations to match fixed targets from successful target-task rollouts.

On LIBERO, we use Action Attack, the targeted action-matching
baseline~\cite{tma} evaluated in TRAP~\cite{trap}.
On SimplerEnv, we evaluate Action Attack, \(\mathrm{TRAP}_{\text{CoT-only}}\),
and TRAP, which align actions, CoT, or both, respectively.
We compare each of these three attacks with its TRACO variant, retaining
the corresponding alignment loss terms.

\paragraph{Implementation details}
Unless otherwise specified, global perturbations use an \(\ell_\infty\)
bound of \(\epsilon=8/255\).
Patches cover approximately \(3.5\%\) and \(4.7\%\)
of the image area on LIBERO and SimplerEnv, respectively.
In the global setting on LIBERO, we report TRACO results
within a budget of 15 rounds, with 30 updates per round.
In the patch setting, the budget is 10 rounds with 100 updates per round
in both environments.
We optimize each offline perturbation for 8,000 gradient updates with a
batch size of 4.
The step sizes are \(1/255\) for LIBERO global perturbations,
\(4/255\) for LIBERO patches, and \(8/255\) for SimplerEnv patches.
We set the PGS position tolerance to \(\kappa=4\) cm on LIBERO and \(\kappa=8\) cm on SimplerEnv.
We optimize each adversarial perturbation on a single NVIDIA A100 GPU
with 80 GB of memory.

\subsection{Overall Performance}
\label{sec:overall-performance}

On LIBERO, TRACO substantially improves task redirection over Action Attack
in the global setting, as shown in Table~\ref{tab:main-results-libero}.
Average Seen ASR increases from \(46.9\%\) to \(85.2\%\).
This advantage extends to Unseen layouts, where ASR increases from
\(43.2\%\) to \(75.9\%\).
Both ASR and Hijack Score improve for all 5 models on both splits.
In the patch setting, TRACO also improves average ASR and Hijack Score on
both splits, with gains varying across models.
Together, these results support using execution feedback to optimize
visual attacks across the three action-generation paradigms.

On SimplerEnv, we examine whether TRACO can improve existing attacks while
retaining their alignment loss terms. As shown in
Table~\ref{tab:instructvla-cot-extension-asr}, TRACO improves both ASR and
Hijack Score for \(\mathrm{TRAP}_{\text{CoT-only}}\) and TRAP on Seen and
Unseen layouts. For TRAP, the strongest offline baseline, ASR rises from
\(64.0\%\) to \(72.8\%\) on Seen layouts and from \(44.0\%\) to \(53.6\%\)
on Unseen layouts. These gains support TRACO as a reusable closed-loop
optimization framework for existing targeted attacks.

\paragraph{Qualitative examples}
Figure~\ref{fig:qualitative-examples} shows the robot completing the
attacker-specified task under the source instruction.
The 3D plots show that the attacked trajectories follow target-task motion
while differing locally from the clean target trajectories.
Such deviations can change the behavior needed to continue the task.
During attack construction, TRACO therefore refreshes target supervision
at the states reached under the current perturbation.

\subsection{Ablation and Mechanism Analysis}
\label{sec:closed-loop-feedback-ablation}
\label{sec:mechanism-analysis}

\paragraph{Component ablation}
 To assess the effects of EDS and PGS, we conduct ablation studies
  on PI in the global and patch settings.
  \textit{w/o} EDS, we keep the initial observations and target supervision fixed.
  \textit{w/o} PGS, we replace progress-guided selection with random sampling using the same number of samples.
\begin{table}[!t]
\centering
\caption{Ablation of EDS and PGS on PI. Without EDS, the initial observations and target supervision remain fixed. Both variants without PGS use random sampling with the same sample counts.}
\label{tab:opts-opr-ablation}
\tablebodyfont
\renewcommand{\arraystretch}{1.0}
\setlength{\tabcolsep}{5.0pt}
\begin{tabular}{@{}cccccc@{}}
\toprule
EDS & PGS & \multicolumn{2}{c}{Seen} & \multicolumn{2}{c}{Unseen} \\
\cmidrule(lr){3-4}\cmidrule(lr){5-6}
& & ASR (\%) $\uparrow$ & \(H\) $\uparrow$ & ASR (\%) $\uparrow$ & \(H\) $\uparrow$ \\
\midrule
\multicolumn{6}{@{}l}{\textbf{Global}} \\
\addlinespace[2pt]
-- & -- & 43.6 & 0.2723 & 35.2 & 0.1671 \\
\(\checkmark\) & -- & 58.4 & 0.4415 & 54.8 & 0.3445 \\
\rowcolor{tracegray}
\(\checkmark\) & \(\checkmark\) & \textbf{89.2} & \textbf{0.6279} & \textbf{79.6} & \textbf{0.5334} \\
\midrule
\multicolumn{6}{@{}l}{\textbf{Patch}} \\
\addlinespace[2pt]
-- & -- & 50.4 & 0.2929 & 49.6 & 0.3113 \\
\(\checkmark\) & -- & 64.8 & 0.3817 & 62.4 & 0.3797 \\
\rowcolor{tracegray}
\(\checkmark\) & \(\checkmark\) & \textbf{69.6} & \textbf{0.4348} & \textbf{67.2} & \textbf{0.4458} \\
\bottomrule
\end{tabular}
\end{table}

As shown in Table~\ref{tab:opts-opr-ablation}, EDS improves ASR and
\(H\) over fixed supervision on Seen and Unseen layouts in both settings.
In the global setting, Seen and Unseen ASR increase by \(14.8\)\% and
\(19.6\)\%, respectively.
EDS rebuilds target supervision at the states reached under the current
perturbation, allowing subsequent updates to follow the evolving
attacked rollout.
These gains support adapting supervision to the states encountered
during execution, where the behavior needed to continue the target
task can change.

PGS complements EDS by organizing this refreshed supervision around
task progress.
Full TRACO achieves higher ASR and \(H\) than EDS alone in both settings.
In the global setting, Seen and Unseen ASR increase by \(30.8\)\% and
\(24.8\)\%, respectively.
In the patch setting, ASR increases by \(4.8\)\% on
both splits.
These improvements suggest that combining frontier and prefix
supervision helps the robot complete the target task.
Frontier supervision encourages further progress, while prefix
supervision helps retain the behavior needed to reach the frontier.

\begin{figure}[!b]
  \centering
  \includegraphics[width=\linewidth]{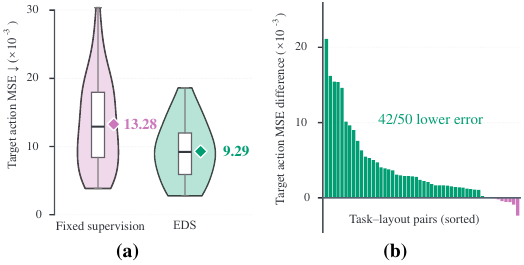}
  \caption{Effect of supervision refresh on target action error.
  Results cover five tasks with 10 Unseen layouts each.
  For each task--layout pair, we collect one rollout under each
  perturbation and evaluate both perturbations on all recorded observations.
  Normalized-action MSE is averaged within each rollout and then
  equally across the two rollouts.
  (a) Distributions of the 50 mean MSE values for Fixed and EDS.
  Boxes show interquartile ranges and median lines.
  Diamonds mark means.
  (b) Sorted differences in mean MSE for each task--layout pair
  (Fixed $-$ EDS). Positive values indicate lower error with EDS.}
  \label{fig:opts-common-state-distribution}
\end{figure}

\begin{figure*}[!t]
  \centering
  \includegraphics[width=\textwidth]{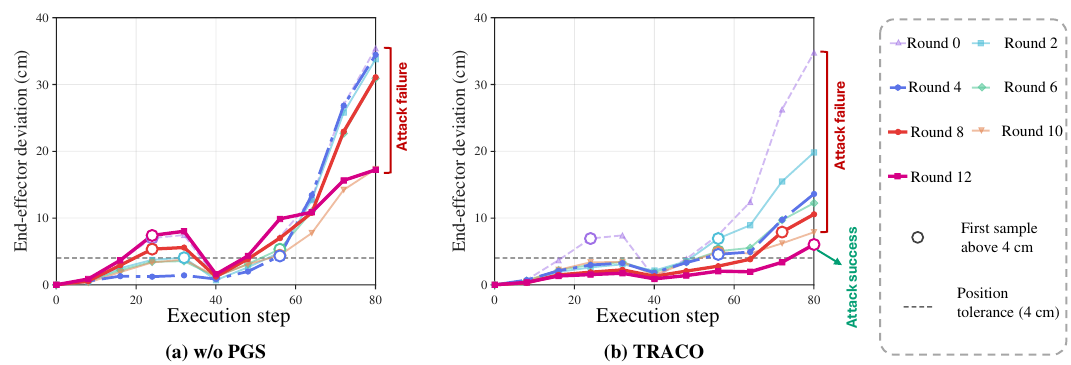}
  \caption{End-effector position error with and without PGS.
  Each curve shows end-effector distances to the successful target
  reference for one optimization round.
  Panels (a) and (b) compare \textit{w/o} PGS and TRACO on PI
  in the global setting.
  Open circles mark the first sampled position error above
  the 4~cm tolerance.
  Attack success denotes target-task completion.}
  \label{fig:opr-progress}
\end{figure*}

\begin{figure}[!t]
  \centering
  \includegraphics[width=\linewidth]{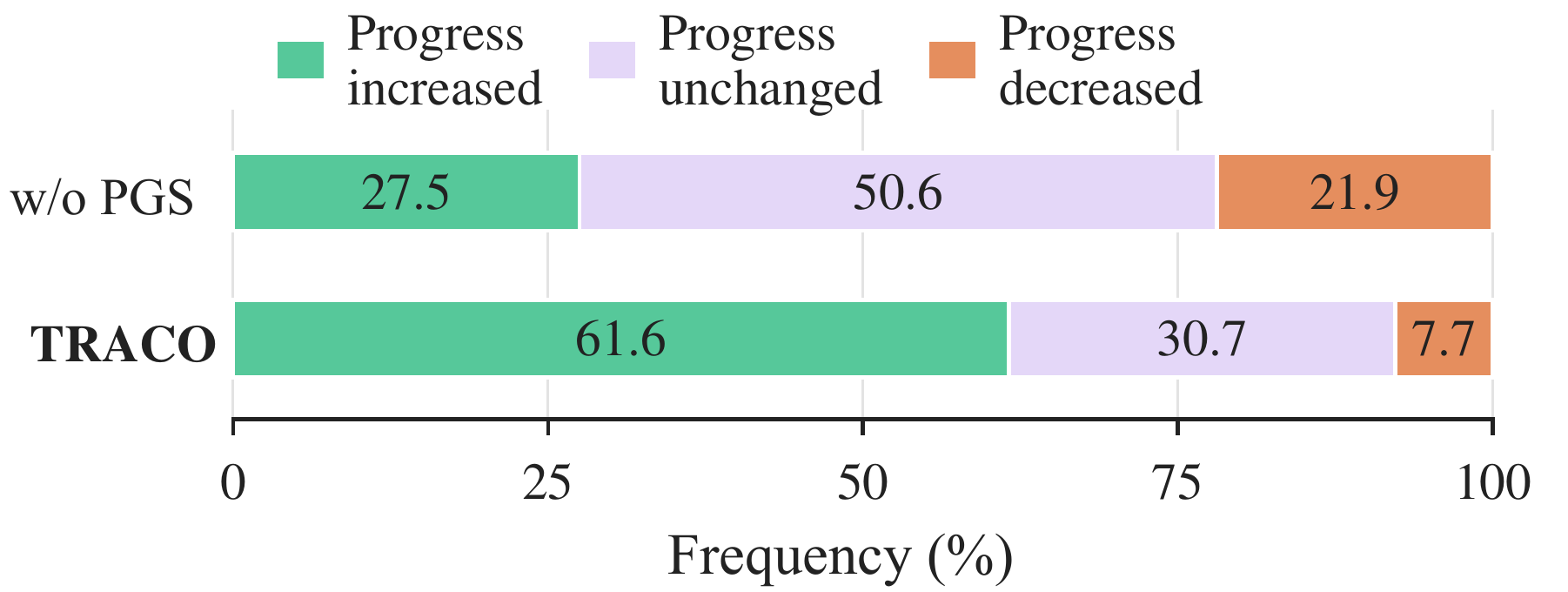}
    \caption{Progress gains and regressions with and without PGS.
  Bars show the mean percentages of progress increases, no changes,
  and decreases between consecutive optimization rounds.
  Progress is measured using the passing prefix under the
  position tolerance (4~cm), with target-task completion assigned full progress.
  Results are averaged across five tasks on PI in the global setting.}
  \label{fig:opr-update-outcomes}
\end{figure}

\paragraph{Effect of EDS on target action error}
We examine whether EDS reduces target action error at states reached
during attacked execution.
On PI in the global setting, we compare perturbations optimized using
fixed supervision or EDS, with PGS disabled in both cases.
We collect rollouts under both perturbations across 5 tasks with
10 Unseen layouts each.
Both perturbations are then evaluated on every recorded observation.
For each unperturbed observation, we query the VLA under the target
instruction to obtain a shared reference action.
We then compute the MSE between this reference and the source-instruction
action predicted under each perturbation.
As shown in Fig.~\ref{fig:opts-common-state-distribution}, EDS reduces
mean target action MSE by \(30.0\%\), with lower error in 42 of 50
task--layout pairs.
The lower error on shared states is consistent with EDS rebuilding
supervision as the attacked rollout evolves, allowing subsequent
updates to follow target behavior at newly reached states.

\paragraph{Effect of PGS on task progress}
We next examine whether PGS helps advance task progress while preserving
earlier gains.
On PI in the global setting, we compare TRACO with \textit{w/o} PGS,
with EDS retained in both.
Figure~\ref{fig:opr-progress}(a,b) shows rollouts from the same
initial condition under perturbations obtained at different
optimization rounds.
Each curve plots the end-effector distance to a successful target
reference at corresponding rollout steps.
A later first violation of the position tolerance indicates
a longer passing prefix.
\textit{w/o} PGS, further updates can move the first sampled violation
to an earlier rollout step, indicating a loss of previously
achieved progress.
In contrast, TRACO extends the passing prefix to later steps,
and its round-12 rollout completes the target task.

To assess whether progress accumulates across tasks, we compare
consecutive-round rollouts from the same initial condition.
We record whether progress increases, remains unchanged, or decreases
after each round of optimization.
The outcome frequencies are averaged over 10 construction initial
conditions per task and then over 5 tasks.
As shown in Fig.~\ref{fig:opr-update-outcomes}, adding PGS increases
the frequency of progress advances from \(27.5\%\) to \(61.6\%\),
while reducing the frequency of regressions from \(21.9\%\) to \(7.7\%\).
These results suggest that focusing supervision on the frontier while
continuing to supervise earlier passing states helps task progress
accumulate across rounds.

\subsection{Additional Evaluations}
\label{sec:additional-evaluations}

\paragraph{Current-round supervision and historical aggregation}
DAgger~\cite{ross2011dagger} trains a policy on expert-labeled
states accumulated across iterations.
TRACO instead keeps the VLA frozen and rebuilds supervision
from current attacked rollouts to optimize the visual perturbation.
To examine whether historical aggregation benefits this optimization,
we compare current-round sampling with DAgger-style historical
aggregation on PI in the global setting.
As shown in Table~\ref{tab:opts-history}, current-round supervision
improves Seen and Unseen ASR by \(22.0\)\% and \(24.8\)\%, 
respectively, with higher \(H\) on both splits.
Earlier perturbations may lead to states that the current
perturbation no longer reaches.
Focusing on current-round states keeps supervision tied to
the execution being optimized.

\begin{table}[!b]
\centering
\caption{Current-round and historical supervision on PI. Both variants disable PGS and use the same number of samples per update.}
\label{tab:opts-history}
\begingroup
\definecolor{tracehighlight}{HTML}{B45A5A}
\tablebodyfont
\renewcommand{\arraystretch}{1.15}
\setlength{\tabcolsep}{2pt}
\begin{tabular*}{\linewidth}{@{\extracolsep{\fill}}lrrrr@{}}
\toprule
\multirow{2}{*}{\textbf{Supervision source}} & \multicolumn{2}{c}{Seen} & \multicolumn{2}{c}{Unseen} \\
\cmidrule(lr){2-3}\cmidrule(l){4-5}
 & \multicolumn{1}{c}{ASR (\%)\(\uparrow\)} & \multicolumn{1}{c}{\(H\uparrow\)} & \multicolumn{1}{c}{ASR (\%)\(\uparrow\)} & \multicolumn{1}{c}{\(H\uparrow\)} \\
\midrule
Historical sampling & 36.4 & 0.2094 & 30.0 & 0.1100 \\
Current round only & \textcolor{tracehighlight}{\textbf{58.4}} & \textcolor{tracehighlight}{\textbf{0.4415}} & \textcolor{tracehighlight}{\textbf{54.8}} & \textcolor{tracehighlight}{\textbf{0.3445}} \\
\bottomrule
\end{tabular*}
\endgroup

\end{table}

\begin{figure*}[!t]
  \centering
  \includegraphics[width=\textwidth]{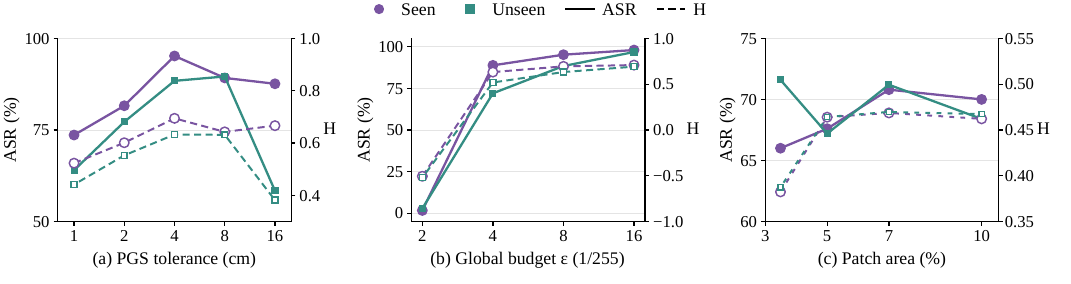}
  \caption{Parameter sensitivity of TRACO on PI.
  (a) PGS position tolerance. (b) Global perturbation budget. (c) Patch area. All three studies use flow-time probes at 0.50 and 0.75, as detailed in Appendix~\ref{sec:flow-probe-sensitivity}.}
  \label{fig:trace-sensitivity}
\end{figure*}

\begin{figure*}[t]
  \centering
  \includegraphics[width=\textwidth]{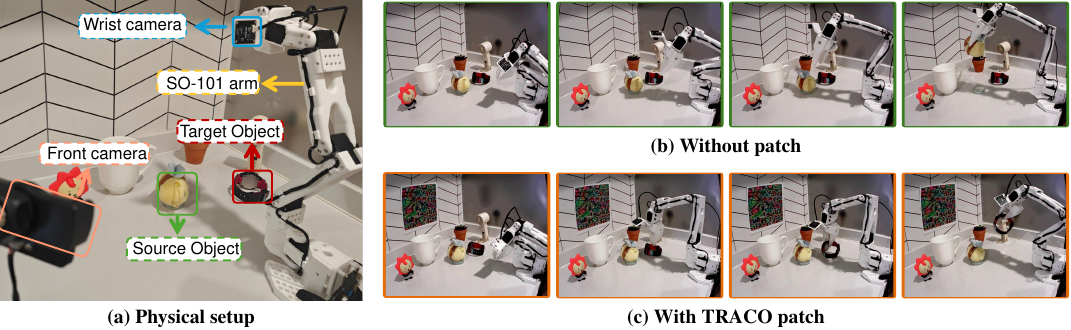}
  \caption{Physical setup and real-world task redirection with a printed TRACO patch.
  Both executions use the same source instruction.}
  \label{fig:real-robot-overview}
\end{figure*}

\paragraph{Parameter sensitivity}
To examine how the position tolerance \(\kappa\) affects task redirection, we vary it on PI and report five-task averages in Fig.~\ref{fig:trace-sensitivity}(a).
Increasing the PGS position tolerance from 1 to 4 cm improves both ASR and \(H\),
whereas a tolerance of 16 cm sharply reduces Unseen performance. The tolerance
determines where the passing prefix ends and where PGS focuses supervision.
A tight tolerance can stop the prefix at small positional differences,
while a loose tolerance can extend it despite substantial deviations from
the reference. These results favor a moderate tolerance that allows trajectory
variation while still identifying substantial deviations.
Figure~\ref{fig:trace-sensitivity}(b) shows higher ASR and \(H\) at larger
global perturbation budgets. For patches, Fig.~\ref{fig:trace-sensitivity}(c)
shows that increasing the area from \(5\%\) to \(7\%\) improves both metrics.

\begin{table}[!t]
\centering
\caption{Task redirection under instruction variations.}
\label{tab:instruction-variation}
\begingroup
\definecolor{tracehighlight}{HTML}{B45A5A}
\fontsize{8.5}{10.5}\selectfont
\renewcommand{\arraystretch}{1.08}
\setlength{\tabcolsep}{1pt}
\begin{tabular*}{\linewidth}{@{\extracolsep{\fill}}ll*{3}{r}@{}}
\toprule
& & \multicolumn{3}{c}{ASR (\%) / \(H\)} \\
\cmidrule(lr){3-5}
Method & Split & \multicolumn{1}{c}{Original} & \multicolumn{1}{c}{Paraphrased} & \multicolumn{1}{c}{Extra-Context} \\
\midrule
\multicolumn{5}{l}{\textbf{LIBERO Global \(\cdot\) Five-model average}} \\
\multirow{2}{*}{\shortstack[l]{Action\\Attack}} & Seen & \({46.9}/{0.2342}\) & \({48.5}/{0.2672}\) & \({45.4}/{0.2325}\) \\
 & Unseen & \({43.2}/{0.1758}\) & \({45.8}/{0.2127}\) & \({41.4}/{0.1701}\) \\
\multirow{2}{*}{\textbf{TRACO}} & Seen & \({\textcolor{tracehighlight}{\mathbf{85.2}}}/{\textcolor{tracehighlight}{\mathbf{0.5887}}}\) & \({\textcolor{tracehighlight}{\mathbf{81.2}}}/{\textcolor{tracehighlight}{\mathbf{0.5954}}}\) & \({\textcolor{tracehighlight}{\mathbf{85.6}}}/{\textcolor{tracehighlight}{\mathbf{0.6041}}}\) \\
 & Unseen & \({75.9}/{0.4819}\) & \({76.0}/{0.5288}\) & \({77.0}/{0.5082}\) \\
\midrule
\multicolumn{5}{l}{\textbf{SimplerEnv Patch \(\cdot\) InstructVLA}} \\
\multirow{2}{*}{\shortstack[l]{Action\\Attack}} & Seen & \({8.8}/{-0.4456}\) & \({10.8}/{-0.4195}\) & \({7.2}/{-0.4494}\) \\
 & Unseen & \({9.6}/{-0.5474}\) & \({9.6}/{-0.5056}\) & \({8.0}/{-0.5144}\) \\
\multirow{2}{*}{+\ \textbf{TRACO}} & Seen & \({\textcolor{tracehighlight}{\mathbf{12.8}}}/{\textcolor{tracehighlight}{\mathbf{-0.3809}}}\) & \({\textcolor{tracehighlight}{\mathbf{11.6}}}/{\textcolor{tracehighlight}{\mathbf{-0.3321}}}\) & \({\textcolor{tracehighlight}{\mathbf{9.2}}}/{\textcolor{tracehighlight}{\mathbf{-0.3635}}}\) \\
 & Unseen & \({7.2}/{-0.4800}\) & \({9.6}/{-0.4103}\) & \({8.4}/{-0.4168}\) \\
\cmidrule(lr){1-5}
\multirow{2}{*}{TRAP} & Seen & \({64.0}/{0.4831}\) & \({24.4}/{0.0013}\) & \({16.8}/{-0.0578}\) \\
 & Unseen & \({44.0}/{0.2663}\) & \({\textcolor{tracehighlight}{\mathbf{24.8}}}/{-0.0169}\) & \({\textcolor{tracehighlight}{\mathbf{18.8}}}/{-0.1251}\) \\
\multirow{2}{*}{+\ \textbf{TRACO}} & Seen & \({\textcolor{tracehighlight}{\mathbf{72.8}}}/{\textcolor{tracehighlight}{\mathbf{0.5293}}}\) & \({19.6}/{\textcolor{tracehighlight}{\mathbf{0.0461}}}\) & \({16.4}/{\textcolor{tracehighlight}{\mathbf{0.0533}}}\) \\
 & Unseen & \({53.6}/{0.2747}\) & \({17.2}/{-0.1607}\) & \({13.2}/{-0.2028}\) \\
\bottomrule
\end{tabular*}
\endgroup
\end{table}

\newcommand{\transferMaxSeen}{12.0}
\newcommand{\transferMaxUnseen}{8.0}
\newcommand{\transferCleanSeen}{93.6}
\newcommand{\transferCleanUnseen}{90.4}

\begin{table}[!t]
\centering
\caption{Transferability of adversarial patches across InstructVLA variants.}
\label{tab:instructvla-transfer}
\begingroup
\definecolor{tracehighlight}{HTML}{B45A5A}
\tablebodyfont
\renewcommand{\arraystretch}{1.15}
\setlength{\tabcolsep}{2pt}
\begin{tabular*}{\linewidth}{@{\extracolsep{\fill}}w{l}{80pt}w{r}{37pt}w{r}{37pt}w{r}{37pt}w{r}{37pt}@{}}
\toprule
\multirow{2}{*}{\textbf{Method}} & \multicolumn{2}{c}{Seen} & \multicolumn{2}{c}{Unseen} \\
\cmidrule(lr){2-3}\cmidrule(l){4-5}
 & \multicolumn{1}{c}{ASR (\%)\(\uparrow\)} & \multicolumn{1}{c}{\(H\uparrow\)} & \multicolumn{1}{c}{ASR (\%)\(\uparrow\)} & \multicolumn{1}{c}{\(H\uparrow\)} \\
\midrule
\(\mathrm{TRAP}_{\text{CoT-only}}\) & 0.8 & \(-\)0.4393 & 2.4 & \(-\)0.4706 \\
\hspace{5pt}+\ \textbf{TRACO} & \textcolor{tracehighlight}{\textbf{3.2}} & \textcolor{tracehighlight}{\textbf{\(-\)0.3579}} & \textcolor{tracehighlight}{\textbf{3.2}} & \textcolor{tracehighlight}{\textbf{\(-\)0.3324}} \\
\midrule
TRAP & 9.6 & \(-\)0.3847 & 7.2 & \(-\)0.3855 \\
\hspace{5pt}+\ \textbf{TRACO} & \textcolor{tracehighlight}{\textbf{12.0}} & \textcolor{tracehighlight}{\textbf{\(-\)0.2358}} & \textcolor{tracehighlight}{\textbf{8.0}} & \textcolor{tracehighlight}{\textbf{\(-\)0.3433}} \\
\bottomrule
\end{tabular*}
\endgroup

\end{table}

\paragraph{Instruction variation}
Following TRAP~\cite{trap}, we evaluate source-instruction paraphrases and
extra context while keeping each perturbation fixed. We average results over
  two variants for each modified instruction condition, with examples in
  Table~\ref{tab:source-instruction-variants}. As shown in Table~\ref{tab:instruction-variation}, TRACO largely
preserves its average ASR across the 5 models in the LIBERO global setting.
In contrast, ASR drops sharply for TRAP with and without TRACO
on InstructVLA in the SimplerEnv patch setting. During attack construction, TRACO refreshes supervision
as execution states change, but still optimizes the perturbation under the
original source instruction. Changing the wording may therefore alter the induced behavior even
when the intended task remains the same.

\paragraph{Transfer across policy variants}
We keep the patches learned on CoT-enabled InstructVLA fixed and evaluate
them on a variant with proprioceptive inputs and no CoT.
As shown in Table~\ref{tab:instructvla-transfer},
  transfer ASR reaches at most \(12.0\%\) on Seen layouts
  and \(8.0\%\) on Unseen layouts. The negative Hijack Scores indicate that transferred
trajectories remain closer to source behavior on average.
For both CoT-only and joint attacks, TRACO yields higher ASR and Hijack Score
than the corresponding offline attacks on both splits.

\subsection{Real-World Evaluation}
\label{sec:real-robot-pilot}

We evaluate printed-patch attacks on StarVLA's GR00T policy adapted to the
SO-101 arm, using the setup shown in Figure~\ref{fig:real-robot-overview}(a).
The source task is to pick up the toy bee, while the attack target is to
pick up the watch. To improve physical robustness, we incorporate
homography transformations and color calibration into patch optimization,
following TRAP~\cite{trap}. Details are provided in Appendix~\ref{sec:physical-patch-calibration}.
Across 15 trials per condition, the clean
policy achieves \(100.0\%\) and \(93.3\%\) success on the source and target
tasks, respectively. With patches printed and placed at the same wall
location, TRACO achieves \(53.3\%\) ASR (\(8/15\)), compared with
\(26.7\%\) (\(4/15\)) for Action Attack. As shown in
Figure~\ref{fig:real-robot-overview}(b,c), the printed TRACO patch changes which
object the robot picks up while the source instruction remains unchanged.
During patch construction, TRACO refreshes target supervision at the states
reached under the current patch, so optimization can respond to changes in
the robot's trajectory. The higher physical ASR suggests that closed-loop optimization can strengthen 
printed-patch attacks on real robots.

\subsection{Construction Cost and Limitations}
\label{sec:construction-cost-limitations}

\begin{table}[!t]
\centering
\caption{Average attack construction cost per task on LIBERO and SimplerEnv. Times include model loading and auxiliary processing.}
\label{tab:construction-cost-empirical}
\begingroup
\definecolor{tracehighlight}{HTML}{B45A5A}
\tablebodyfont
\renewcommand{\arraystretch}{1.15}
\setlength{\tabcolsep}{0pt}
\noindent\begin{tabular}{@{}>{\raggedright\arraybackslash}m{92pt}@{\hspace{4pt}}>{\raggedleft\arraybackslash}m{36pt}@{\hspace{3pt}}>{\raggedleft\arraybackslash}m{36pt}@{\hspace{6pt}}>{\raggedleft\arraybackslash}m{36pt}@{\hspace{3pt}}>{\raggedleft\arraybackslash}m{36pt}@{}}
\toprule
\multirow{2}{*}{Metric} & \multicolumn{2}{c}{\shortstack{\textbf{LIBERO Global}\\(PI)}} & \multicolumn{2}{c}{\shortstack{\textbf{SimplerEnv Patch}\\(InstructVLA)}} \\
\cmidrule(lr){2-3}\cmidrule(l){4-5}
 & \multicolumn{1}{c}{\shortstack{Action\\Attack}} & \multicolumn{1}{c}{\textbf{TRACO}} & \multicolumn{1}{c}{TRAP} & \multicolumn{1}{c}{\textbf{TRACO}} \\
\midrule
\textbf{Total time (min)} & \(\approx\!396.3\) & \textcolor{tracehighlight}{\(\approx\!\mathbf{88.8}\)} & \(\approx\!513.4\) & \textcolor{tracehighlight}{\(\approx\!\mathbf{353.1}\)} \\
Optimization time (min) & \(\approx\!396.3\) & \(\approx\!50.0\) & \(\approx\!513.4\) & \(\approx\!280.8\) \\
Rollout time (min) & \(0.0\) & \(\approx\!38.8\) & \(0.0\) & \(\approx\!72.3\) \\
\midrule
\textbf{Sample uses (\(\times 10^3\))} & \(\approx\!32\) & \textcolor{tracehighlight}{\(\mathbf{\approx\!6.1}\)} & \(\approx\!32\) & \textcolor{tracehighlight}{\(\mathbf{\approx\!19.1}\)} \\
\bottomrule
\end{tabular}
\endgroup

\end{table}

\paragraph{Construction cost}
Table~\ref{tab:construction-cost-empirical} compares construction costs averaged
over 5 tasks under each method's optimization settings.
TRACO takes less total time and processes fewer samples across gradient
updates than the offline
baselines in both environments.
Across all layouts, TRACO selects an average of 45.6 records per round on
LIBERO and 20.6 on SimplerEnv, with maxima of 94 and 31, respectively.
Optimization accounts for most of its construction time.
The additional rollout stage covers closed-loop execution and supervision
construction. TRACO collects an average of 50 and 56 rollouts per task in
the two environments.
These executions provide the state and progress feedback that lets TRACO
refresh target supervision and focus subsequent updates while
maintaining supervision on earlier behavior.

However, TRACO requires repeated interaction with the environment during
attack construction, adding execution time and computational cost.
The benefit of supervision refresh also depends on whether the frozen VLA
can guide the robot toward the target task from attack-induced states.
When the policy struggles to recover, querying it at the current state may
still provide ineffective guidance.
PGS estimates progress from end-effector positions, which do not fully
capture whether object interactions such as grasping succeed.

\section{Conclusion}
\label{sec:conclusion}

We introduced TRACO, a closed-loop optimization framework for visual
task-redirection attacks. It rebuilds target supervision at the states reached
under attack, enabling existing attacks to use execution feedback while
retaining their alignment objectives. Within this framework, PGS selects states
for supervision to advance the target task while preserving earlier behavior.
Experiments in simulation and on a real robot demonstrate the effectiveness
of TRACO for visual task redirection. Together, these findings highlight the
security implications of closing the loop between visual attack optimization
and execution.

\FloatBarrier
\appendix
\subsection{Target Alignment Details}
\label{sec:architecture-loss-details}

We instantiate the alignment loss in Eq.~\eqref{eq:target-alignment-loss}
for sequence, continuous-action, and flow-field targets. At each selected
observation \(o_i\), the source prediction and target supervision use the
conditions in Eq.~\eqref{eq:on-policy-target-evaluations}.

\paragraph{Sequence targets}
For action tokens or CoT, the target query supplies a sequence
\(\mathbf z_i^{\mathrm{tgt}}=(z_{i,1}^{\mathrm{tgt}},\ldots,
z_{i,M_i}^{\mathrm{tgt}})\). We minimize its length-normalized negative
log-likelihood with teacher forcing:
\begin{equation}
\label{eq:token-sequence-loss}
\ell_i(\delta)=-\frac{1}{M_i}\sum_{h=1}^{M_i}
\log p_\theta\!\left(z_{i,h}^{\mathrm{tgt}}
\mid z_{i,<h}^{\mathrm{tgt}},g_\delta(o_i),\csrc\right).
\end{equation}
Here, \(M_i\) is the sequence length, \(z_{i,<h}^{\mathrm{tgt}}\) contains
the preceding target tokens, and \(p_\theta\) is the frozen VLA's next-token
distribution.

\paragraph{Continuous actions}
For a policy that directly predicts normalized action chunks, we match its
prediction to the target action chunk
\(\mathbf a_i^{\mathrm{tgt}}=\pi_\theta(o_i,\ctgt)\):
\begin{equation}
\label{eq:action-chunk-loss}
\ell_i(\delta)=\operatorname{MSE}\!\left(
\pi_\theta(g_\delta(o_i),\csrc),\mathbf a_i^{\mathrm{tgt}}\right).
\end{equation}
The MSE averages squared differences over all entries of the normalized
action chunks.

\paragraph{Flow-matching policies}
A flow-matching policy generates an action chunk by integrating a
conditional vector field from noise \(\mathbf x(0)\sim p_0\):
\begin{equation}
\label{eq:flow-matching-policy}
\frac{\mathrm d\mathbf x(\tau)}{\mathrm d\tau}
=v_\theta(\mathbf x(\tau),\tau\mid o,c),
\qquad \pi_\theta(o,c)=\mathbf x(1).
\end{equation}
At each selected observation, TRACO records the source-conditioned flow
states \(\mathbf x_i(\tau)\) under \((g_{\delta_r}(o_i),\csrc)\).
At selected flow times \(\tau\in\mathcal T\), it queries the target field
at the same flow state. The alignment loss is
\begin{equation}
\label{eq:field-loss}
\begin{aligned}
\ell_i(\delta)&=\frac{1}{|\mathcal T|}\sum_{\tau\in\mathcal T}
\operatorname{MSE}\!\Bigl(
v_\theta(\mathbf x_i(\tau),\tau\mid g_\delta(o_i),\csrc),\\
&\hspace{43mm}v_i^{\mathrm{tgt}}(\tau)\Bigr),\\
v_i^{\mathrm{tgt}}(\tau)&=
v_\theta(\mathbf x_i(\tau),\tau\mid o_i,\ctgt).
\end{aligned}
\end{equation}
The probe set \(\mathcal T\) selects times on the native solver grid.
Within a round, the recorded flow states and target fields remain fixed,
so updating \(\delta\) changes only the source field at these points.
The next rollout provides the states used to rebuild both the flow probes
and target fields.

\paragraph{Joint objectives}
On InstructVLA, we retain TRAP's action and CoT alignment terms~\cite{trap}.
The action term compares the trajectories represented by predicted action
chunks, while the CoT term uses target-sequence likelihood.
Their joint loss is \(\ell_{\mathrm{CoT}}+2\ell_{\mathrm{act}}\).

\subsection{Sensitivity to Flow-Time Probes}
\label{sec:flow-probe-sensitivity}

We vary the flow-time probe set \(\mathcal T\) for vector-field alignment
on PI across the five LIBERO global task pairs.
As shown in Table~\ref{tab:flow-probe-sensitivity}, the choice of flow times
affects performance, but adding probes does not consistently improve ASR.
The set \(\{0.50,0.75\}\) yields the highest Seen ASR, while
\(\{0.50,0.75,1.00\}\) performs best on Unseen layouts.

For perturbations on LIBERO, we use
\(\mathcal{T}=\{0.25,0.50,0.75\}\) for flow-matching models.
The parameter sensitivity experiments in Fig.~\ref{fig:trace-sensitivity}
use a fixed probe set \(\mathcal{T}=\{0.50,0.75\}\).

\begin{table}[!t]
\centering
\caption{Flow-time probe sensitivity on PI in the LIBERO global setting.
Results are averaged over five tasks.}
\label{tab:flow-probe-sensitivity}
\begingroup
\tablebodyfont
\renewcommand{\arraystretch}{1.12}
\setlength{\tabcolsep}{2pt}
\begin{tabular*}{\linewidth}{@{\extracolsep{\fill}}lrrrr@{}}
\toprule
\multirow{2}{*}{\textbf{Alignment}} & \multicolumn{2}{c}{Seen} & \multicolumn{2}{c}{Unseen} \\
\cmidrule(lr){2-3}\cmidrule(l){4-5}
 & \multicolumn{1}{c}{ASR (\%)\(\uparrow\)} & \multicolumn{1}{c}{\(H\uparrow\)} & \multicolumn{1}{c}{ASR (\%)\(\uparrow\)} & \multicolumn{1}{c}{\(H\uparrow\)} \\
\midrule
\(\mathcal T=\{0.25\}\) & 88.0 & 0.6392 & 83.6 & 0.5969 \\
\(\mathcal T=\{0.50\}\) & 92.4 & 0.6588 & 84.4 & 0.5812 \\
\(\mathcal T=\{0.75\}\) & 91.6 & 0.6702 & 82.0 & 0.5790 \\
\(\mathcal T=\{0.25,0.50\}\) & 87.2 & 0.6330 & 78.4 & 0.5631 \\
\(\mathcal T=\{0.25,0.75\}\) & 93.6 & 0.6845 & 84.0 & 0.5830 \\
\(\mathcal T=\{0.50,0.75\}\) & 95.2 & 0.6943 & 88.4 & 0.6324 \\
\(\mathcal T=\{0.25,0.50,0.75\}\) & 89.2 & 0.6279 & 79.6 & 0.5334 \\
\(\mathcal T=\{0.50,0.75,1.00\}\) & 93.2 & 0.6728 & 90.8 & 0.6416 \\
\bottomrule
\end{tabular*}
\endgroup
\end{table}

\begin{table}[!t]
\centering
\caption{Task pairs in LIBERO and SimplerEnv.}
\label{tab:task-pairs-and-variants}
\begingroup
\tablebodyfont
\renewcommand{\arraystretch}{1.04}
\setlength{\tabcolsep}{3pt}
\begin{tabular}{@{}>{\raggedright\arraybackslash}p{\dimexpr.5\linewidth-\tabcolsep\relax}>{\raggedright\arraybackslash}p{\dimexpr.5\linewidth-\tabcolsep\relax}@{}}
\toprule
\textbf{Source task} & \textbf{Target task} \\
\midrule
\multicolumn{2}{@{}l}{\textit{LIBERO: Global}} \\
Push plate to stove front & Open middle cabinet drawer \\
Open top drawer and place bowl inside & Place wine bottle on cabinet top \\
Place cream cheese in bowl & Turn on stove \\
Place bowl on plate & Place bowl on cabinet top \\
Place wine bottle on rack & Place bowl on stove \\
\midrule
\multicolumn{2}{@{}l}{\textit{LIBERO: Patch}} \\
Place BBQ sauce in basket & Place chocolate pudding in basket \\
Place milk in basket & Place tomato sauce in basket \\
Place cream cheese in basket & Place milk in basket \\
Place salad dressing in basket & Place ketchup in basket \\
Place bowl on plate & Place bowl on cabinet top \\
\midrule
\multicolumn{2}{@{}l}{\textit{SimplerEnv: Patch}} \\
Pick up orange & Pick up Coke can \\
Pick up apple & Pick up orange \\
Pick up apple & Pick up Pepsi can \\
Pick up Coke can & Pick up 7up can \\
Pick up orange & Pick up sponge \\
\bottomrule
\end{tabular}
\endgroup

\end{table}

\begin{table}[!t]
\centering
\caption{Examples of source-instruction variants.}
\label{tab:source-instruction-variants}
\begingroup
\tablebodyfont
\renewcommand{\arraystretch}{1.04}
\setlength{\tabcolsep}{3pt}
\begin{tabular}{@{}>{\raggedright\arraybackslash}p{\dimexpr.23\linewidth-\tabcolsep\relax}>{\raggedright\arraybackslash}p{\dimexpr.77\linewidth-\tabcolsep\relax}@{}}
\toprule
\textbf{Condition} & \textbf{Instruction} \\
\midrule
\multicolumn{2}{@{}l}{\textit{LIBERO: Global}} \\
Original & push the plate to the front of the stove \\
Paraphrased & slide the plate to a position in front of the stove \\
Extra-Context & There are several objects in the workspace. Push the plate to the front of the stove. \\
\midrule
\multicolumn{2}{@{}l}{\textit{SimplerEnv: Patch}} \\
Original & pick orange \\
Paraphrased & Grasp and lift the orange. \\
Extra-Context & There are several objects on the table. Pick the orange. \\
\bottomrule
\end{tabular}
\endgroup

\end{table}

\subsection{Physical Robustness}
\label{sec:physical-patch-calibration}

Following TRAP~\cite{trap}, we account for geometric projection and changes
in patch appearance during physical patch construction.

\begin{figure}[!t]
  \centering
  \includegraphics[width=\columnwidth]{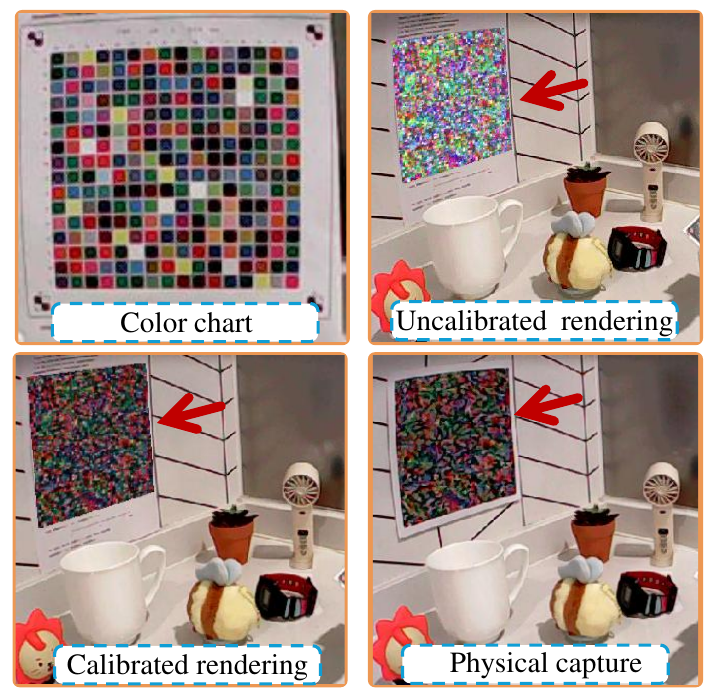}
  \caption{Color calibration for physical patch deployment.}
  \label{fig:real-robot-patch-threeway}
\end{figure}

\paragraph{Homography transformations}
To model the patch's perspective appearance in the camera image, we use a
planar homography. A pixel at \((u,v)\) in the
canonical patch plane maps to image coordinates \((x,y)\) as
\[
\lambda
\begin{bmatrix}
x\\y\\1
\end{bmatrix}
=
\mathbf H
\begin{bmatrix}
u\\v\\1
\end{bmatrix}
=
\begin{bmatrix}
h_{11}&h_{12}&h_{13}\\
h_{21}&h_{22}&h_{23}\\
h_{31}&h_{32}&h_{33}
\end{bmatrix}
\begin{bmatrix}
u\\v\\1
\end{bmatrix},
\]
where \(\mathbf H\) is the homography matrix and \(\lambda\) is the
homogeneous scale factor. We use this mapping to render the patch in the
camera view during optimization.

\paragraph{Color calibration}
For physical deployment, we calibrate colors using a printed reference
chart photographed by the robot's camera. A third-order polynomial mapping
models the color changes introduced by printing and camera capture. We
keep this mapping fixed during optimization to predict the patch's camera
appearance. Figure~\ref{fig:real-robot-patch-threeway} illustrates the
calibration.

\paragraph{Expectation over transformations}
We apply expectation over transformations (EoT)~\cite{athalye2018robust} during patch
optimization. Random brightness changes and Gaussian noise model variation
in illumination and image capture.

\subsection{VLA Model Details}
\label{sec:vla-model-details}

Table~\ref{tab:vla-model-details} lists the VLM backbones and
action-generation interfaces used in our evaluation.
StarVLA~\cite{starvla2026} provides a modular framework that combines
pretrained VLM backbones with different action-generation modules.
Our LIBERO policies use this framework.

\paragraph{PI}
PI follows the \(\pi_0\) design~\cite{pi02024}, coupling a VLM with a
flow-matching action expert. The expert attends to hidden states from
multiple VLM layers to generate continuous action chunks.

\paragraph{GR00T}
GR00T~\cite{bjorck2025gr00t} uses a dual-system architecture. Its VLM encodes the scene and
instruction, and a diffusion-transformer (DiT) action module uses these
features to generate continuous actions through flow matching.

\paragraph{FAST}
FAST uses the FAST tokenizer~\cite{pertsch2025fast} to encode action chunks
as discrete tokens. The VLM predicts these tokens autoregressively, and
the tokenizer decodes them into continuous actions.

\paragraph{OFT}
OFT follows OpenVLA-OFT~\cite{kim2025oft}, using a lightweight multilayer
perceptron to predict continuous actions. This head reads hidden states
at action-token positions and predicts an action chunk in parallel.

\paragraph{LangForce}
LangForce~\cite{langforce2026} uses learnable latent action queries to
connect the VLM to a flow-matching action head. Its training objective
rewards latent actions that help predict the instruction from the scene,
encouraging instruction-dependent behavior.

\paragraph{InstructVLA}
InstructVLA-Generalist~\cite{yang2026instructvla} produces CoT reasoning 
before generating latent action representations.
A flow-matching action expert uses these representations and visual
features to produce continuous actions.

\begin{table}[t]
\centering
\caption{VLM backbones and action interfaces of the evaluated policies.}
\label{tab:vla-model-details}
\begingroup
\tablebodyfont
\setlength{\tabcolsep}{4pt}
\begin{tabular}{@{}>{\raggedright\arraybackslash}p{0.21\columnwidth}>{\raggedright\arraybackslash}p{0.40\columnwidth}>{\raggedright\arraybackslash}p{0.31\columnwidth}@{}}
\toprule
Model & VLM backbone & Action generation \\
\midrule
PI & Qwen3-VL-4B-Instruct & Flow matching \\
GR00T & Qwen3-VL-4B-Instruct & Flow matching \\
FAST & Qwen2.5-VL-3B-Instruct & Autoregressive tokens \\
OFT & Qwen3-VL-4B-Instruct & Direct regression \\
LangForce & Qwen3-VL-4B-Instruct & Flow matching \\
InstructVLA & Eagle2-2B & Flow matching \\
\bottomrule
\end{tabular}
\endgroup
\end{table}

\FloatBarrier

\bibliography{references}
\bibliographystyle{IEEEtran}

\end{document}